%% file: main.tex
\documentclass[10pt,twocolumn,letterpaper]{article}

\usepackage{cvpr}              % To produce the CAMERA-READY version
\usepackage[accsupp]{axessibility}

\input{preamble}

\definecolor{cvprblue}{rgb}{0.21,0.49,0.74}
\usepackage[pagebackref,breaklinks,colorlinks,allcolors=cvprblue]{hyperref}

\def\paperID{10102} % *** Enter the Paper ID here
\def\confName{CVPR}
\def\confYear{2025}

\title{Chat2SVG: Vector Graphics Generation with Large Language Models and Image Diffusion Models}

\author{Ronghuan Wu\\
City University of Hong Kong\\
{\tt\small rh.wu@my.cityu.edu.hk}
\and
Wanchao Su\\
Monash University\\
{\tt\small wanchao.su@monash.edu}
\and
Jing Liao\footnotemark[1]\\
City University of Hong Kong\\
{\tt\small jingliao@cityu.edu.hk}
}

\begin{document}
\begin{sloppypar}  % avoid overfull lines
% \maketitle

\twocolumn[{
\renewcommand\twocolumn[1][]{#1}
\maketitle
    \vspace{-2.0em}
    \setlength\tabcolsep{0.5pt}
    \centering
    \includegraphics[width=0.95\textwidth]{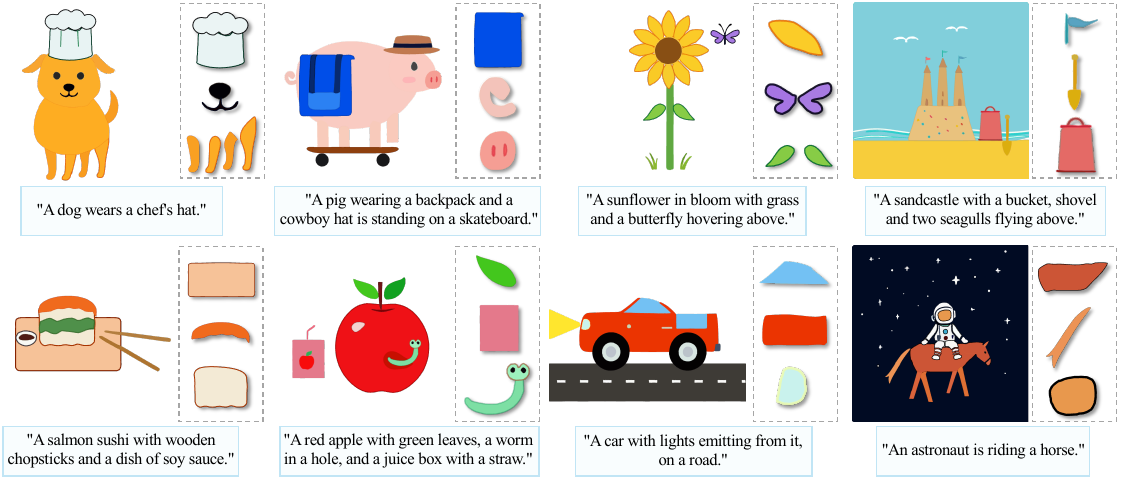}
    \captionsetup{type=figure} % fixes warning "Package caption Warning: The option `hypcap=true' will be ignored for this particular \caption"
    \captionof{figure}{SVG examples generated by our Chat2SVG. We highlight some shapes to demonstrate semantic clarity and path quality.}
    \vspace{0.5em}
    \label{figure:teaser}
}]
{
  \renewcommand{\thefootnote}%
    {\fnsymbol{footnote}}
  \footnotetext[1]{Corresponding Author.}
}

\input{chat2svg}
{
    \small
    \bibliographystyle{ieeenat_fullname}
    \bibliography{main}
}

\input{suppl}

\end{sloppypar}
\end{document}

%% file: preamble.tex
\newcommand{\sysName}{Chat2SVG}

\usepackage[table]{xcolor}
\usepackage{array, adjustbox}
\usepackage{longtable}
\usepackage{listings}
\lstdefinelanguage{XML}
{
  morestring=[b]",
  morestring=[s]{>}{<},
  morecomment=[s]{<?}{?>},
  stringstyle=\color{black},
  identifierstyle=\color{black},
  keywordstyle=\color{black},
  morekeywords={xmlns,id,viewBox,version,type,cx,cy,rx,ry,x,y,width,height,fill,points,stroke,stroke-width,d}
}

%% file: chat2svg.tex
\begin{abstract}
Scalable Vector Graphics (SVG) has become the de facto standard for vector graphics in digital design, offering resolution independence and precise control over individual elements. Despite their advantages, creating high-quality SVG content remains challenging, as it demands technical expertise with professional editing software and a considerable time investment to craft complex shapes. Recent text-to-SVG generation methods aim to make vector graphics creation more accessible, but they still encounter limitations in shape regularity, generalization ability, and expressiveness. To address these challenges, we introduce \sysName, a hybrid framework that combines the strengths of Large Language Models (LLMs) and image diffusion models for text-to-SVG generation. Our approach first uses an LLM to generate semantically meaningful SVG templates from basic geometric primitives. Guided by image diffusion models, a dual-stage optimization pipeline refines paths in latent space and adjusts point coordinates to enhance geometric complexity. Extensive experiments show that \sysName~outperforms existing methods in visual fidelity, path regularity, and semantic alignment. Additionally, our system enables intuitive editing through natural language instructions, making professional vector graphics creation accessible to all users. Our code is available at \url{https://chat2svg.github.io/}.
\end{abstract}

%===========================================
% INTRODUCTION
%===========================================
\section{Introduction}
\label{sec:intro}

Scalable Vector Graphics (SVG), a vector image format based on geometric shapes, has become the standard for modern digital design, offering resolution independence and precise control over individual elements.
However, creating high-quality SVG content remains challenging for non-expert users, as it requires both expertise with professional design software and considerable time investment to create complex shapes.
To make vector graphics creation more accessible, recent research has focused on developing text-to-SVG systems that enable users to express their creative ideas through simple text prompts rather than complex manual editing.

Existing approaches have explored \textit{image-based} methods for text-to-SVG generation, iteratively optimizing a large collection of shape elements (\eg, cubic B\'{e}zier curves, typically $100$ to $1000$) by rendering them into images with differentiable rasterizers~\cite{Li2020DiffVG} and evaluating them using text-image similarity metrics like CLIP loss~\cite{radford2021learning} and Score Distillation Sampling (SDS) loss~\cite{poole2022dreamfusion}.
While these image-based methods~\cite{frans2021clipdraw, jain2022vectorfusion,xing2023diffsketcher,xing2024svgdreamer,zhang2024text} can generate visually impressive SVG through the combination of numerous paths and powerful image models, they face a critical limitation in maintaining the regularity and semantics of shapes. Specifically, semantic components that should be represented as single elements often end up fragmented across multiple overlapping paths.
Although these fragmented paths, when viewed collectively, can yield visually appealing outputs, they fundamentally conflict with professional design principles, where each semantic component is intentionally crafted as a single, regularized path.

Given SVG is defined using Extensible Markup Language (XML), another solution to text-to-SVG synthesis involves \textit{language-based} methods, which treat SVG scripts as text input.
Recent works~\cite{wu2023iconshop,tang2024strokenuwa} have proposed specialized tokenization strategies to embed SVG scripts and trained Sequence-To-Sequence (seq2seq) models on domain-specific datasets (\eg, icons and fonts).
While these approaches achieve good generation quality within their training domains, they suffer from limited generalization due to the absence of large-scale, general-purpose text-SVG training data.
Meanwhile, some works~\cite{openai2023gpt4, sharma2024vision} show that while state-of-the-art Large Language Models (LLMs) can generate basic geometric shapes (\eg, circles and rectangles) with layouts matching text prompts, they struggle to produce the complex geometric details required for professional SVG applications. Consequently, the limited generalization and poor expressiveness of language-based methods hinder their adoption for text-to-SVG generation.

To address the aforementioned problems, we propose a novel hybrid framework that leverages the complementary strengths of \textit{image-based} and \textit{language-based} methods. Our approach first utilizes an LLM to synthesize SVG templates composed of basic primitives and then optimizes their geometric details guided by image diffusion models. This hybrid framework addresses the key limitations of both paradigms: The LLM-based template generation overcomes the domain-specific generalization constraints of traditional language models while ensuring shape regularity, as each element naturally corresponds to a single semantic component. The subsequent image-based optimization then enhances the expressiveness of these well-structured templates by capturing complex geometric details that LLMs alone struggle to produce.
Specifically, to fully exploit LLMs' ability to create reasonable and complex layouts that match text prompts, we design an SVG-oriented prompt with multiple stages, including prompt expansion, SVG script generation, and visual refinement.
To refine the generated SVG templates, we implement a dual-stage optimization strategy: (1) We first use an image diffusion model to synthesize detailed image-level targets, then follow~\cite{zhang2024text} to optimize path-level latent vectors with a pretrained SVG VAE, eliminating common issues such as self-intersections and jagged curves;
(2) We further refine geometric details by directly optimizing the point coordinates to capture fine-grained visual elements.
Comprehensive experiments show that \sysName~consistently outperforms existing methods regarding overall visual quality, individual path regularity, and semantic coherence. 
Furthermore, our system enables intuitive editing through iterative natural language instructions, making vector graphics creation more accessible to non-expert users. Our contributions are summarized as follows:
\begin{itemize}
    \item \textbf{Hybrid SVG Generation Framework}. We introduce a novel text-to-SVG generation paradigm that combines Large Language Models with image diffusion models to produce high-quality SVG outputs.
    
    \item \textbf{SVG-Oriented Prompt Design}. We develop a specialized prompt system that directs LLMs to generate SVG templates using basic geometric primitives.
    
    \item \textbf{Dual-Stage Optimization}. We implement a two-phase optimization process that preserves the semantic meaning of each shape while eliminating artifacts such as self-intersecting and jagged paths.
    
    \item \textbf{Iterative Editing}. We enable iterative refinement of SVG through natural language instructions, making SVG creation more accessible.
\end{itemize}

%===========================================
% RELATED WORK
%===========================================
\section{Related Work}
\label{sec:related}

\subsection{Large Language Models for Design}
LLMs have emerged as powerful tools for graphics design tasks, demonstrating remarkable capabilities in understanding and reasoning about complex design specifications across diverse domains.
Recent works have explored their applications in (1) visual layout and composition, including poster designs~\cite{tang2023layoutnuwa, cheng2024graphic, yang2024posterllava, lin2024layoutprompter}, 3D scene arrangements~\cite{wang2024chat2layout, hu2024scenecraft, tam2024scenemotifcoder, zhang2023scenewiz3d, sun20233d, aguina2024open, fu2025anyhome}, and placement of basic shapes for image generation~\cite{feng2024layoutgpt, yamada2024l3go};
(2) content creation and manipulation for shapes~\cite{ganeshan2024parsel}, materials~\cite{huang2024blenderalchemy}, and animation~\cite{liu2024logomotion, tseng2024keyframer}; and (3) design understanding~\cite{kulits2024re, wu2024gpt}. While these demonstrate LLMs' versatility, their potential for vector graphics remains unexplored. Our work addresses this gap by showing that LLMs can effectively generate structured SVG templates, extending their capabilities to the domain of vector graphics creation.

\subsection{Text-Guided Vector Graphics Generation}
Text-to-SVG generation approaches can be roughly categorized into \textit{image-based} and \textit{language-based} methods.
Image-based methods~\cite{frans2021clipdraw, jain2022vectorfusion,xing2023diffsketcher,xing2024svgdreamer,zhang2024text} start with a large collection of randomly initialized shapes, render vector graphics with a differentiable renderer~\cite{Li2020DiffVG}, and iteratively optimize path parameters by minimizing text-image similarity losses (\eg, CLIP~\cite{radford2021learning} loss and Score Distillation Sampling~\cite{poole2022dreamfusion} loss).
Despite their visually pleasant appearance, these methods often generate SVG results containing multiple fragmented paths that lack individual semantic correspondence, which is undesirable for real design scenarios.
Language-based methods~\cite{wu2023iconshop, belouadi2023automatikz, tang2024strokenuwa} design specialized tokenization strategies to encode vector graphics into discrete tokens, which are then concatenated with text tokens, and train seq2seq models like autoregressive transformers~\cite{vaswani2017attention} on domain-specific datasets (\eg, icons and fonts).
Although these approaches are conceptually elegant, their generalizability is limited by the scarcity of large-scale, general-purpose vector graphics datasets.
Some works~\cite{openai2023gpt4, sharma2024vision} have explored LLMs' vector graphics generation capabilities, showing that while LLMs can create reasonable layouts matching text prompts, they struggle to produce complex geometric shapes.
Our \sysName~combines image-based and language-based methods to address the limitations of each paradigm, producing semantically meaningful SVG where each path corresponds to a distinct visual element and contains refined geometric details.

\subsection{Vector Graphics Representation Learning}
Vector graphics representation learning is essential for sketch-based retrieval, reconstruction, and generation tasks.
A seminal study, SketchRNN~\cite{ha2017neural}, combines a Recurrent Neural Network (RNN) and a VAE to learn vector representations.
\citet{lopes2019learned} trained a VAE to represent image-level font styles, then passed the latent embedding into a decoder for vector font generation.
To increase reconstruction quality, later approaches like Sketchformer~\cite{ribeiro2020sketchformer} and DeepSVG~\cite{carlier2020deepsvg} adopted Transformer~\cite{vaswani2017attention} architectures, followed by dual-modality methods~\cite{wang2021deepvecfont, liu2023dualvector, wang2023deepvecfont} that leveraged both vector and image features.
However, encoding the entire complex SVG into a single latent embedding often causes detail loss. To address this limitation, \citet{zhang2024text} proposed path-level representational learning, creating a smooth latent space for individual paths. We leverage these path-level latent embeddings in our optimization to eliminate self-intersecting and jagged paths.

%===========================================
% METHOD
%===========================================

\begin{figure*}[t]
    \centering
    \includegraphics[width=0.83\linewidth]{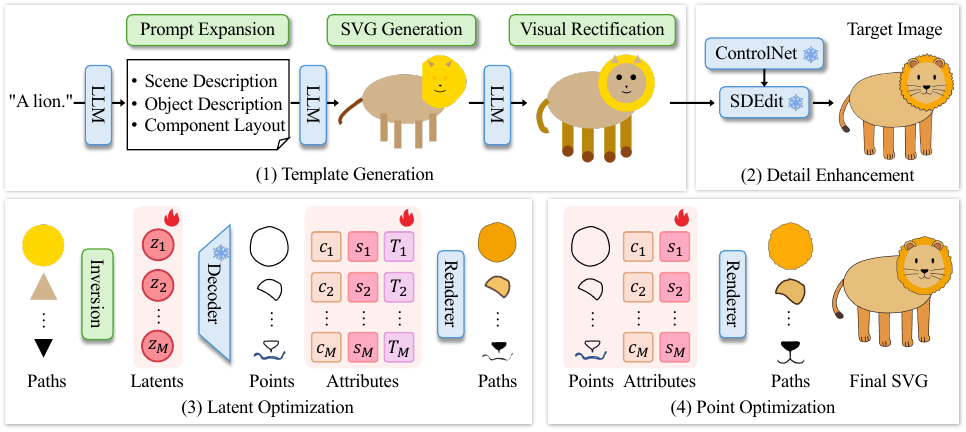}
    \caption{%
      The system pipeline of \sysName.
      Given a text prompt, our system first leverages an LLM to generate an SVG template composed of basic geometric primitives. The rendered template is enhanced through SDEdit~\cite{meng2021sdedit} with ControlNet~\cite{zhang2023adding} to add visual details while preserving the overall composition, yielding a target image. The SVG then undergoes a dual-stage optimization process to match the target image.
        (1) Primitives are converted to latent embeddings through latent inversion and optimized along with their visual attributes (\ie, filling colors $c_i$, stroke properties $s_i$, and transformation matrices $T_i$).
        (2) Point-level optimization is performed to refine the geometric details of SVG paths.
    }
    \label{fig:pipeline}
\end{figure*}

\section{Method}
\label{sec:method}

Our \sysName, as shown in Figure~\ref{fig:pipeline}, begins with an SVG-oriented prompting approach (Section~\ref{sec:template_generation}) that guides an LLM to create reasonable SVG templates.
We then perform a dual-stage optimization (Section~\ref{sec:optimization}) to enhance the geometric details of SVG templates by optimizing SVG paths in both latent and control point space, guided by image diffusion models.

\subsection{SVG Template Generation with LLMs}
\label{sec:template_generation}

\textbf{Prompt Expansion}.
Users often provide vague and brief prompts to the LLM when describing their desired graphics.
Such unstructured input may lead to oversimplified SVG (regarding both the number of elements and structures) that fails to reflect the intended design.
Consequently, we propose a three-level prompt expansion strategy that systematically delineates the initial prompts:
(1) Scene-level: Starting with an initial prompt, we instruct the LLM to analyze it holistically and identify essential objects that should appear. The LLM then expands the prompt by suggesting complementary objects to enhance scene completeness.
(2) Object-level: For each object in the expanded prompt, we guide the LLM to systematically break it down into its components. For instance, when describing a lion, the LLM deconstructs it into distinct parts such as the body, head, mane, legs, eyes, ears, and tail. This decomposition ensures that no critical components are missing during SVG script generation.
(3) Layout-level: After object decomposition, we direct the LLM to develop a comprehensive layout plan. This includes determining the position and size of each element on the canvas, selecting appropriate colors for visual harmony, and establishing clear spatial relationships to ensure a cohesive overall composition.

\noindent{\textbf{SVG Script Generation}.}
After prompt expansion, we obtain a detailed scene description.
We then convert this natural language specification into SVG scripts in XML format.
Since the LLM has limitations in synthesizing geometrically complex paths, we constrain the shapes to a carefully selected set of basic primitives: rectangles, ellipses, lines, polylines, polygons, and short paths.
We set the canvas size to $512 \times 512$.
Each geometric element is assigned a unique ID and includes semantic annotations in its comments.

\noindent{\textbf{Visual Rectification}.}
Since prompt expansion is purely text-based, even when the generated SVG script accurately follows the detailed prompt, visual inconsistencies (\eg, misaligned components, disproportionate scaling, and incorrect path ordering) can emerge during rendering.
For example, in Figure~\ref{fig:pipeline}, the initial SVG template shows a lion with a misshapen mane and missing facial features.
We thus adopt a visual rectification strategy in which we render the SVG and provide the rendered image back to the LLM (with vision capability) for inspection.
The LLM analyzes the visual output, identifies any inconsistencies or oddities, and generates corrected SVG code.
This visual refinement loop can be performed iteratively.
In our experiments, we find that two iterations of refinement are typically sufficient to generate well-structured SVG templates with appropriate spatial layouts.

Furthermore, to enhance the quality of prompt expansion and SVG script generation, we provide curated in-context examples in the prompts.
The complete set of prompts and examples is available in the supplementary material.

\subsection{SVG Optimization Guided by Image Diffusion}
\label{sec:optimization}

\textbf{Detail Enhancement}.
While the primitive shapes in the SVG template have accurate semantic meaning, they lack the geometric details necessary for a professional design.
To improve visual expressiveness, we first render the SVG template into a raster image, and then use an image editing method to generate a more detailed version that serves as our optimization target.
Specifically, we employ SDEdit~\cite{meng2021sdedit}, which enhances the input image by first adding noise and then progressively denoising it with an image diffusion model to produce outputs with richer details, such as the contour of the lion's mane illustrated in Figure~\ref{fig:pipeline}.
To maintain structural similarity between the enhanced output and the original image, we incorporate a ControlNet (tile version)~\cite{zhang2023adding} into our pipeline. By using Gaussian-blurred versions of the initial images as control signals, the ControlNet effectively maintains the overall composition throughout the enhancement process.

Apart from geometric details, this image-to-image translation process also introduces new semantic parts such as the lion's beard. To faithfully reproduce the target image, we propose using the Segment Anything Model (SAM)~\cite{kirillov2023segment} to identify these new decorative details.
Specifically, we denote the rendered images of the initial SVG template and its diffusion-enhanced counterpart as $I_{\mathrm{template}}$ and $I_{\mathrm{target}}$, respectively. We apply SAM to extract mask sets from both images: $\{m_i\}_{i=1}^Q$ for the template image and $\{m_j\}_{j=1}^K$ for the target image, where $Q$ and $K$ represent the number of masks in each collection.
Our approach is based on the observation that the diffusion process simultaneously generates fine decorative details that are absent in the original template (\eg, the beard of the lion in Figure~\ref{fig:pipeline}) while preserving the structural integrity of larger components (such as the body). Consequently, we implement an area-based mask-filtering strategy to identify these newly generated elements.
For each mask $m_j$ detected in $I_{\mathrm{target}}$, we evaluate it against all masks in $\{m_i\}_{i=1}^Q$ using Intersection over Union (IoU) metrics~\cite{rezatofighi2019generalized}. When $m_j$ overlaps with a template mask $m_i$ but their IoU falls below a threshold, we classify $m_j$ as a new decorative component.
Alternatively, if their IoU exceeds the threshold, we interpret $m_j$ as a variation of an existing component.
For all filtered masks classified as new decorative shapes, we approximate their boundaries with polygons and integrate them into the SVG script.

% \noindent{\textbf{Notation}}.
We define an SVG script $G$ as a collection of $M$ paths (\ie, shapes), $G=\{P_i\}_{i=1}^M$. Each path $P_i$ consists of a sequence of $N_i$ commands, $P_i=\{C_i^j\}_{j=1}^{N_i}$. A command $C_i^j=(U_i^j, V_i^j)$ is defined by its type $U_i^j\in\{\texttt{M}, \texttt{C}\}$ and its associated control points $V_i^j$, where $\texttt{M}$ represents \texttt{Move} and $\texttt{C}$ represents cubic B\'{e}zier curves.
For consistency, we convert all other primitive shapes (\eg, rectangles and ellipses) in the SVG template into cubic Bézier curves.

Previous image-based methods~\cite{frans2021clipdraw, jain2022vectorfusion, xing2023diffsketcher, xing2024svgdreamer} optimize SVG shapes by directly manipulating the control points $V_i^j$.
Despite convenience, this approach often leads to artifacts such as self-intersecting curves and unnatural deformations.
To overcome these limitations, \citet{zhang2024text} introduced a path-level SVG VAE. This model's latent space effectively captures common shape patterns and geometric constraints, allowing for the optimization of path latent vectors at a higher level to produce smooth outputs.

\noindent{\textbf{Latent Optimization}.}
We leverage this pretrained SVG VAE to conduct the latent optimization.
However, their SVG encoder expects a fixed number of commands ($10$ cubic B\'{e}zier curves), while our SVG primitives contain varying (typically fewer) commands. 
To resolve this mismatch, we develop a latent inversion process that converts our primitives into latent embeddings.
We start with a randomly sampled latent vector $z$ and decode it into a command sequence. Then we evenly sample points $X = \{x_i\}_{i=1}^{N}$ along its contour. Similarly, we sample an equal number of points $Y = \{y_j\}_{j=1}^{N}$ along the contour of our target primitive shape. Using these point sets, we compute the Earth Mover's Distance (EMD)~\cite{achlioptas2018learning, fan2017point, yu2018pu}:
\begin{equation}
    \ell_{\mathrm{EMD}}(X, Y) = \min_{\phi: X\rightarrow Y} \sum_{x\in X} \left\| x-\phi(x) \right\|_2,
\end{equation}
where $\phi$ represents a bijective mapping between the point sets. By minimizing this EMD loss, we gradually optimize the latent embedding $z$ to match the target primitive.

After obtaining latent vectors $\{z_i\}_{i=1}^M$ for all paths, we jointly optimize these vectors along with visual attribute parameters: filling colors $\{c_i\}_{i=1}^M$, stroke properties (color and width) $\{s_i\}_{i=1}^M$, and transformation matrices $\{T_i\}_{i=1}^M$.
At each optimization iteration $t$, we decode the latent vectors into command sequences and apply transformation matrices to the control points. We then combine color and stroke attributes together, and use a differentiable rasterizer~\cite{Li2020DiffVG} to render the complete SVG into an image $I_t$.
Our optimization objective consists of three loss terms:
(1) An MSE loss $\ell_{\mathrm{MSE}}$ between the rendered image $I_t$ and target image $I_{\mathrm{target}}$ to ensure visual similarity;
(2) A curvature loss that reduces sharp bends and fluctuations along the contour by computing the discrete second derivative:
\begin{equation}
    \ell_{\mathrm{curvature}} = \frac{\sum\limits_{i=1}^M \sum\limits_{j=1}^{N_i-2} \left\| V_i^{j} - 2V_i^{j+1} + V_i^{j+2} \right\|_2^2}{\sum\limits_{i=1}^M (N_i-2)};
\end{equation}
(3) A path-level IoU loss that preserves the overall composition and prevents paths from drifting too far from their initial positions:
\begin{equation}
    \ell_{\mathrm{IoU}}=\frac{1}{M}\sum_{i=1}^M \left(1-\frac{|m_i^t \cap m_i^0|}{|m_i^t \cup m_i^0|}\right),
\end{equation}
where $m_i^t$ and $m_i^0$ denote the binary masks of the $i$-th path at the current iteration $t$ and its initial state, respectively.

The final loss function combines these three terms with empirically determined weights:
\begin{equation}
    \ell_{\mathrm{latent}}=\ell_{\mathrm{MSE}} + \lambda_1 \ell_{\mathrm{curvature}} + \lambda_2 \ell_{\mathrm{IoU}},
\end{equation}
where $\lambda_1=5e-4$ and $\lambda_2=5e-6$.

\noindent{\textbf{Point Optimization}.}
While latent optimization effectively aligns SVG shapes with their target positions and contours, it has a notable limitation: the resulting shapes tend to be overly smooth, lacking the intricate details that are often essential in professional vector graphics. For example, in Figure~\ref{fig:pipeline}, the lion's mane after latent optimization does not capture the fine contour present in the target image.
Therefore, we add a second optimization stage that directly refines the control points $V_i^j$ of the paths, similar to previous image-based works~\cite{frans2021clipdraw, jain2022vectorfusion, xing2023diffsketcher, xing2024svgdreamer}.
To achieve finer granularity in shape control, we split each cubic B\'{e}zier curve at its midpoint, effectively doubling the number of control points in the SVG. This increased control point density allows for more precise shape adjustments.
The point optimization stage employs a loss function that combines the MSE loss and the curvature loss:
\begin{equation}
    \ell_{\mathrm{point}}=\ell_{\mathrm{MSE}} + \lambda_3 \ell_{\mathrm{curvature}}.
\end{equation}
The parameter $\lambda_3$ decreases linearly from $1e-3$ to $5e-5$ throughout the optimization process.

\subsection{Iterative Editing}
During SVG script generation (Section~\ref{sec:template_generation}), we instruct the LLM to annotate the semantic label of each path.
This design enables users to refine the initial SVG template using natural language, as the LLM can precisely locate and modify specific elements based on semantic understanding.

Our system supports iterative refinement through multiple rounds of natural language instructions.
However, when passing the edited SVG template image into SDEdit, image diffusion models may introduce different shape variations and decorative details, which compromise the user's intention to maintain the already optimized shapes.
To preserve consistency across editing iterations, we instruct the LLM to output both the modified SVG script and a list of specifically changed paths.
This precise tracking enables selective optimization of modified shapes while preserving unaltered elements from previous iterations, maintaining visual coherence throughout the entire refinement process.

%===========================================
% EXPERIMENTS
%===========================================

\begin{figure*}[t]
    \centering
    \includegraphics[width=0.9\linewidth]{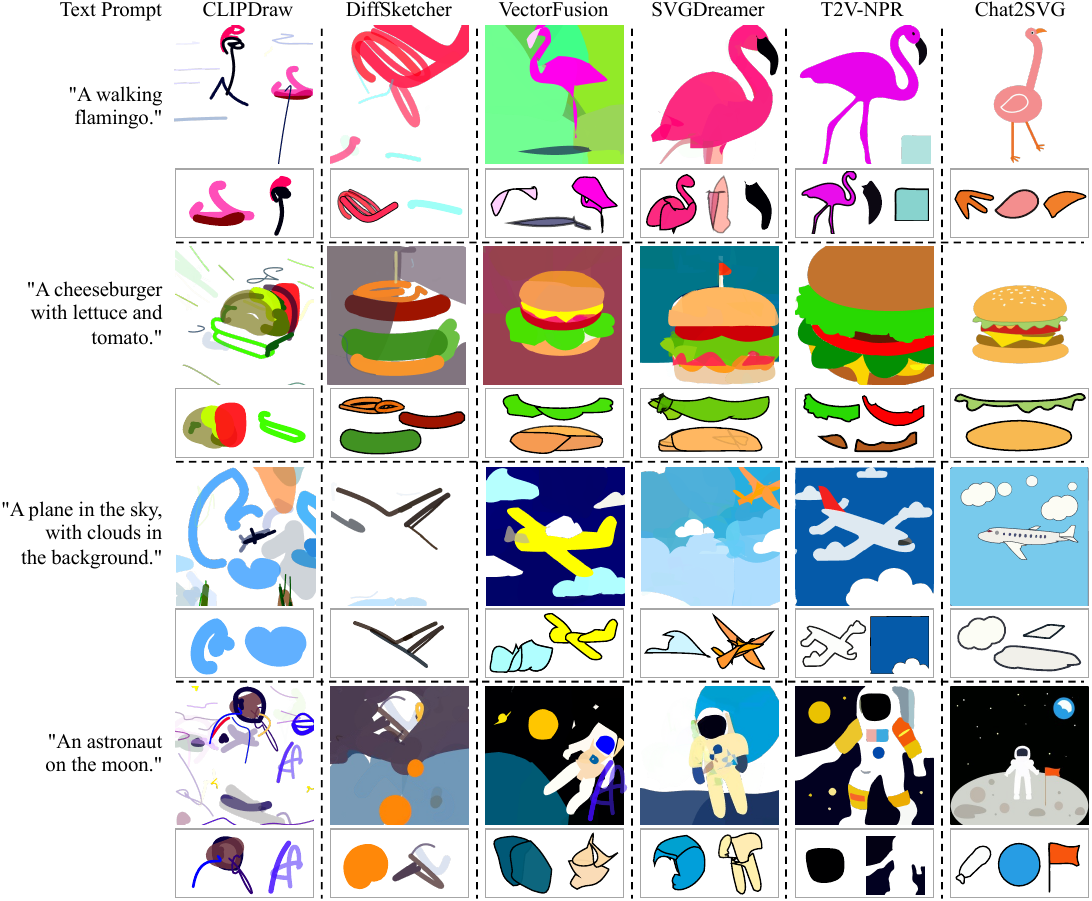}
    \caption{%
        \textbf{Qualitative Comparison}.
        (1) Methods refining open-ended strokes, \ie, CLIPDraw~\cite{frans2021clipdraw} and DiffSketcher~\cite{xing2023diffsketcher}, often produce distorted and disorganized strokes to approximate objects, presenting messy appearance and poor text alignment.
        (2) VectorFusion~\cite{jain2022vectorfusion} and SVGDreamer~\cite{xing2024svgdreamer} produce elements that consist of multiple jagged, irregular, and fragmented shapes, such as the body of the flamingo (first row) and the plane (third row).
        (3) T2V-NPR~\cite{zhang2024text} attempts to resolve these issues by learning a latent representation of paths and merging fragmented shapes. However, it still cannot guarantee the semantic meanings of the paths, leading to less-semantic paths such as a plane body with surrounding clouds in the third row.
        In contrast, our method produces SVG with superior text alignment, higher visual quality, and well-structured paths exhibiting geometric regularity and clear semantic definition.
    }
    \label{fig:comparison}
\end{figure*}

\section{Experiments}
\label{sec:experiment}

\textbf{Implementation Details}.
In our experiments, we use \texttt{claude-3-5-sonnet} as our backbone LLM model due to its leading generation capabilities.
To evaluate our approach, we create a prompt set by having the LLM generate $125$ text prompts across five categories: animals, food, objects, scenes, and novel concept combinations.
For each prompt, we generate one template and perform $2$ rounds of visual rectification. This process is repeated $5$ times, yielding $(1+2)\times 5=15$ candidate SVG templates per prompt.
These templates are rendered as images, and we follow a standard practice to select the highest quality SVG using the ImageReward~\cite{xu2024imagereward} metric.
Detailed optimization configurations are provided in the supplementary materials.

\noindent{\textbf{Baselines}}.
We compare our approach against five state-of-the-art image-based text-to-SVG generation methods: CLIPDraw~\cite{frans2021clipdraw}, DiffSketcher~\cite{xing2023diffsketcher}, VectorFusion~\cite{jain2022vectorfusion}, SVGDreamer~\cite{xing2024svgdreamer}, and T2V-NPR~\cite{zhang2024text}.
Current language-based methods~\cite{wu2023iconshop, tang2024strokenuwa} are confined to specific categories, with no support for general text-to-SVG generation, so we do not include them in our comparison.
For a fair comparison, we set these baseline methods to use a similar number of shapes as our optimized SVG.

\noindent{\textbf{Evaluation Metrics}}.
We evaluate the quality of our generated SVG across three dimensions: visual fidelity, path regularity, and semantic alignment.
\begin{itemize}
    \item \textit{Image-level Fidelity}. We evaluate visual fidelity using a ground-truth dataset of well-designed vector graphics, specifically $52,805$ colored SVG files downloaded from SVGRepo\footnote{\url{https://www.svgrepo.com}}. We compute the Fr\'{e}chet Inception Distance (FID)~\cite{heusel2017gans} between the rendered images of our generated SVG and this professional design collection, using features extracted by the CLIP image encoder~\cite{radford2021learning}.
    
    \item \textit{Path-level Regularity}. We assess path quality using a transformer-based Auto-Encoder trained on the \textit{FIGR-8-SVG}~\cite{clouatre2019figr} dataset, following the DeepSVG~\cite{carlier2020deepsvg} architecture. Our model, trained with reconstruction tasks, encodes each drawing command $C_i^j$ into a latent vector. We represent each path as the mean of its commands' latent embeddings, and calculate the FID between these path representations and the ground truth paths from \textit{FIGR-8-SVG}. This metric quantifies how closely our generated paths align with professional SVG design patterns.
    
    \item \textit{Text-level Alignment}. We assess semantic alignment by computing the CLIP score~\cite{radford2021learning} between the text prompt and the rendered SVG image.
\end{itemize}

\subsection{Comparison with Existing Methods}

\textbf{Quantitative Comparison}.
Table~\ref{tab:quantitative_comparison} shows the evaluation metrics for all baseline methods.
Our method achieves the best image FID score, indicating that our generated SVG closely aligns with professional design patterns.
Regarding path regularity, our method yields paths that are closest to the professionally designed SVG, as evidenced by the lowest path
FID. Furthermore, our \sysName~achieves the highest score in text-SVG alignment, validating the significance of LLM-generated SVG templates.

\begin{table}[t]
    \centering
    \footnotesize
    \begin{tabular}{lccc}
        \toprule
        Method & 
        \begin{tabular}[c]{@{}c@{}}Image\\FID\end{tabular} $\big\downarrow$ & 
        \begin{tabular}[c]{@{}c@{}}Path\\FID\end{tabular} $\big\downarrow$ & 
        \begin{tabular}[c]{@{}c@{}}Text\\Alignment\end{tabular} $\big\uparrow$ \\
        \toprule
        CLIPDraw~\cite{frans2021clipdraw} & $46.77$ & $70.13$ & $0.3048$\\
        DiffSketcher~\cite{xing2023diffsketcher} & $44.89$ & $66.48$ & $0.2623$ \\
        VectorFusion~\cite{jain2022vectorfusion} & $39.52$ & $56.79$ & $0.2982$ \\
        SVGDreamer~\cite{xing2024svgdreamer} & $35.48$ & $47.95$ & $0.2919$ \\
        T2V-NPR~\cite{zhang2024text} & $39.86$ & $42.03$ & $0.3078$ \\
        \hline
        Chat2SVG (Ours) & $\mathbf{33.31}$ & $\mathbf{39.07}$ & $\mathbf{0.3090}$ \\
        \bottomrule
    \end{tabular}
    \caption{Quantitative comparison of text-to-SVG generation methods across image fidelity, vector regularity, and text alignment.
    }
    \label{tab:quantitative_comparison}
\end{table}

\noindent{\textbf{Qualitative Comparison}}.
In Figure~\ref{fig:comparison}, we present a side-by-side comparison between our \sysName~and the baselines.
Methods that optimize open-ended strokes based on CLIP or Diffusion models (\ie, CLIPDraw~\cite{frans2021clipdraw} and DiffSketcher~\cite{xing2023diffsketcher}) produce results with messy visual appearances and poor text alignment.
These methods sometimes fail to synthesize complete objects, such as the flamingo (first row) and the plane (third row) in Figure~\ref{fig:comparison}.
In contrast, our method generates SVG results with clear layouts and greater fidelity to the prompts.

% Image-based methods
Methods that optimize closed shapes via score distillation of diffusion models (\ie, VectorFusion~\cite{jain2022vectorfusion}, SVGDreamer~\cite{xing2024svgdreamer}, and T2V-NPR~\cite{zhang2024text}) can better align with the input text, but still produce fragmented paths with limited semantic meaning. 
For instance, in the plane example (third row of Figure~\ref{fig:comparison}), VectorFusion and SVGDreamer create planes and clouds composed of multiple jagged, irregular, and fragmented shapes, which only appear meaningful when viewed as a whole. This hinders convenient editing by graphic designers.
T2V-NPR~\cite{zhang2024text} addresses the issue of jagged paths using a path VAE (although it can be overly smooth at times) and reduces fragmented shapes by merging shapes with similar colors.
However, its merging operation ignores the semantic meaning of shapes, resulting in the body of the plane being merged with the surrounding clouds.
Additionally, the semantic meaning of the path is sometimes unclear, as exemplified by the cloud in the bottom-right corner, whose contour is actually approximated by the blue background instead of a true cloud shape.
In contrast, our \sysName~adopts a unique approach by using LLMs to generate paths representing semantic components and performing dual-stage optimization based on image diffusion to enhance path expressiveness, thus ensuring both path regularity and visual fidelity of generated SVG.

\begin{figure}[t]
    \centering
    \includegraphics[width=0.8\linewidth]{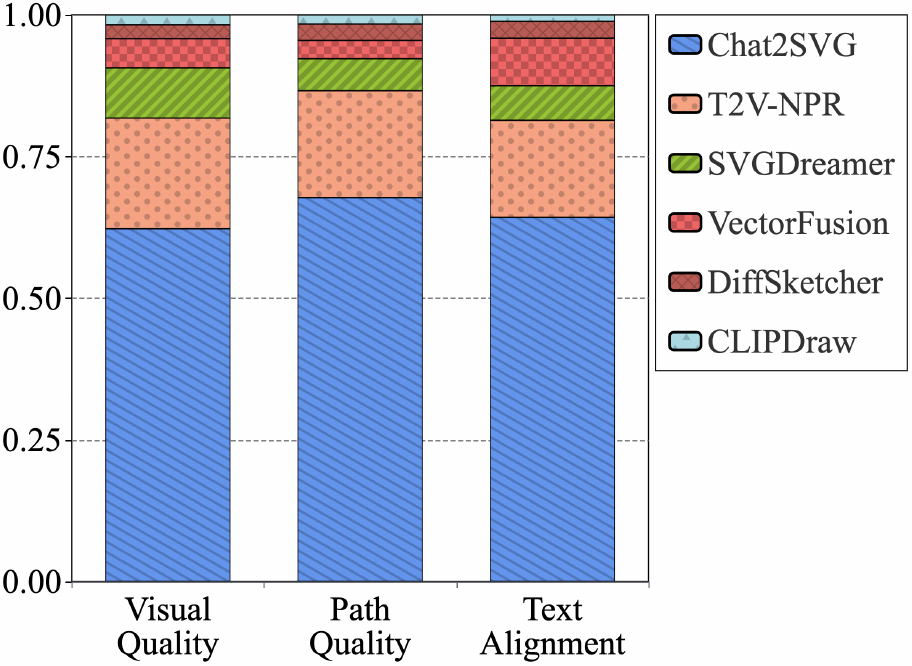}
    \caption{%
        \textbf{User Study}.
        Our \sysName~achieves the highest user selection ratio across all three evaluation criteria.
    }
    \label{fig:user_study}
\end{figure}

\noindent{\textbf{User Study}}.
To further compare \sysName~with baseline methods, we conducted a user study examining the same key aspects: (1) overall visual aesthetics, (2) path regularity, and (3) text-SVG alignment.
We randomly sampled $20$ prompts and generated SVG outputs using our approach and five baseline methods.
For each prompt, participants were shown all generated SVG results along with the corresponding text prompt and highlighted paths.
They were then asked to select the highest quality result regarding the three evaluation criteria.
We recruited $31$ participants ($18$ male, $13$ female) through university mailing lists, with a mean age of $26$ years. Among them, $17$ participants reported prior experience in graphic design, providing a balanced mix of expert and novice perspectives.
Analyzing the average selection ratio across all three metrics (Figure~\ref{fig:user_study}), we found that \sysName-generated results were consistently preferred by participants over baseline methods.

\subsection{Ablation Study}
\begin{figure}[t]
    \centering
    \includegraphics[width=\linewidth]{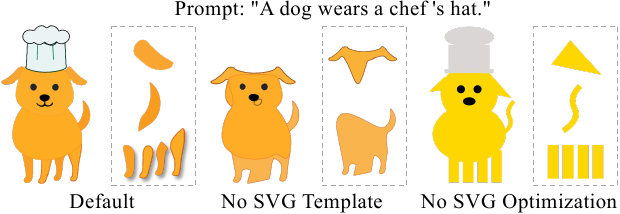}
    \caption{%
        Qualitative results of ablation study.
    }
    \label{fig:ablation}
\end{figure}

We conduct ablation experiments to evaluate the effectiveness of key components in our pipeline (Section~\ref{sec:template_generation}).
First, we remove the SVG template generation and use randomly initialized shapes to approximate target images. As shown in the second row of Table~\ref{tab:ablation}, both image FID and path FID values increase substantially. This performance degradation can be attributed to the substantial gap between randomly initialized shapes and target shapes, making optimization more challenging. Meanwhile, in the second column of Figure~\ref{fig:ablation}, the dog's legs and ears lack semantic-clear components, highlighting the importance of SVG template generation.
Second, we eliminate the dual-stage optimization. As shown in the third row of Table~\ref{tab:ablation}, this results in the lowest path FID, as primitives generated by LLMs naturally align with the path regularity presented in the ground-truth dataset. However, the visual outcome in Figure~\ref{fig:ablation} lacks the necessary details, indicating that the overall visual quality suffers without dual-stage optimization. A more detailed ablation study is provided in the supplementary material.

\begin{table}[t]
    \centering
    \footnotesize
    \begin{tabular}{lccc}
        \toprule
        \sysName~Variant & 
        \begin{tabular}[c]{@{}c@{}}Image\\FID\end{tabular} $\big\downarrow$ & 
        \begin{tabular}[c]{@{}c@{}}Path\\FID\end{tabular} $\big\downarrow$ & 
        \begin{tabular}[c]{@{}c@{}}Text\\Alignment\end{tabular} $\big\uparrow$ \\
        \toprule
        Default & $\mathbf{33.31}$ & $39.07$ & $\mathbf{0.3090}$\\
        No SVG Template & $47.21$ & $70.13$ & $0.2968$\\
        No SVG Optimization & $33.45$ & $\mathbf{36.12}$ & $0.3044$ \\
        \bottomrule
    \end{tabular}
    \caption{
        Qualitative results of the ablation study.
    }
    \label{tab:ablation}
    
\end{table}

\subsection{Iterative Editing Results}
\begin{figure}[t]
    \centering
    \includegraphics[width=0.9\linewidth]{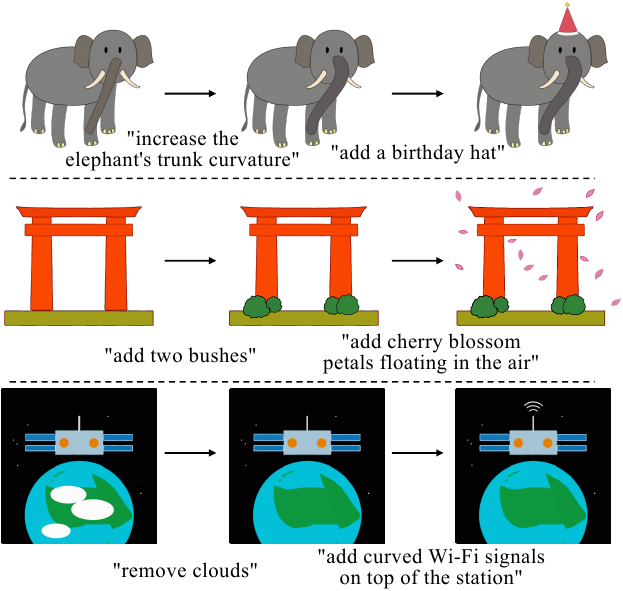}
    \caption{%
        \textbf{Iterative Editing}.
        We perform two rounds of refinement on each SVG template and show the optimized output.
        This figure shows editing types including deletion, modification, and addition.
    }
    \label{fig:iterative_edit}
\end{figure}

In this section, we present the iterative editing results.
As demonstrated in Figure~\ref{fig:iterative_edit}, users can progressively refine the SVG through multiple rounds of natural language instructions.
During editing, we preserve the optimized shapes from previous iterations while only modifying the specified shapes. 
The results showcase how \sysName~can accurately interpret and execute user editing requests, maintaining both semantic coherence and visual quality throughout the modifications.
Additional editing examples are provided in the supplementary material.

%===========================================
% CONCLUSION
%===========================================

\begin{figure}[h]
    \centering
    \includegraphics[width=0.9\linewidth]{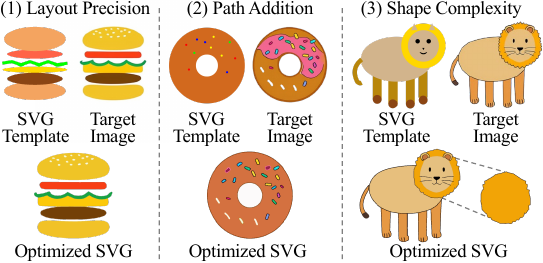}
    \caption{%
        \textbf{Limitations} of our method include imprecise layout generation, missing important visual elements, and insufficient shape complexity.
    }
    \label{fig:limitation}
\end{figure}

\section{Conclusion}
\label{sec:conclusion}

In this paper, we present \sysName, a novel paradigm for text-to-SVG generation that combines LLMs and image diffusion models.
Through our carefully designed SVG template generation and dual-stage SVG optimization pipeline, our method generates high-quality SVG outputs that exhibit strong visual fidelity, path regularity, and text alignment.

Despite our method's superior results, there are several aspects that could be further improved:
(1) LLMs may produce imprecise visual layouts, such as the burger in Figure~\ref{fig:limitation}. This can be alleviated by collecting an SVG dataset and extracting the corresponding layouts to fine-tune LLMs. (2) Our approach uses SAM to identify new parts. However, SAM may overlook important regions, like the pink frosting on the donut in Figure~\ref{fig:limitation}. Incorporating semantic-aware SAM~\cite{li2023semantic} could potentially produce more accurate results.
(3) The pretrained SVG VAE from~\cite{zhang2024text} has a fixed number of commands. This makes it hard to represent intricate shapes. For example, the lion's mane in Figure~\ref{fig:limitation} lacks sufficient detail. Training a new SVG VAE with varying-length commands would help solve this problem.
Our work can be extended to generate other types of vectorized data, such as icons and fonts. Additionally, through consistent editing of SVG across multiple frames, coupled with shape interpolation, our Chat2SVG has the potential to facilitate keyframe animation generation.

\section*{Acknowledgement}
The work described in this paper was fully supported by a GRF grant from the Research Grants Council (RGC) of the Hong Kong Special Administrative Region, China [Project No.  CityU 11216122].

%% file: suppl.tex
\appendix
\clearpage
\setcounter{page}{1}
\maketitlesupplementary

\section{Overview}
This supplementary material provides additional implementation details and experimental results, including:
\begin{itemize}
    \item More details about the optimization procedure (Section~\ref{sec:supp_implement});
    \item Additional ablation studies evaluating the effects of prompt design strategies, SAM-guided path additions, and various optimization stages (Section~\ref{sec:supp_ablation});
    \item Complete LLM prompts used in SVG generation and editing (Section~\ref{sec:supp_prompt});
    \item More generated and edited SVG examples (Section~\ref{sec:supp_results}) are available on the project page.
\end{itemize}

\section{Optimization Details}
\label{sec:supp_implement}
Our SVG optimization consists of three sequential stages: latent inversion, latent optimization, and point optimization. Each stage runs for $500$ iterations.
During the latent optimization stage, we initialize the stroke color to black with a width of $0.8$. We also set the gradient of the RGBA color's alpha channel to zero to forbid transparency optimization.
Upon completing latent optimization, we apply transformation matrices to the points, which allows us to bypass the need for optimizing transformation matrices in the subsequent point optimization stage.
Optimizing an SVG containing approximately $30$ shapes requires around $10$ minutes when executed on a single NVIDIA RTX $4090$ GPU, consuming roughly $5$GB of GPU memory.

\section{Additional Ablation Study}
\label{sec:supp_ablation}
\begin{figure*}[t]
    \centering
    \includegraphics[width=0.95\linewidth]{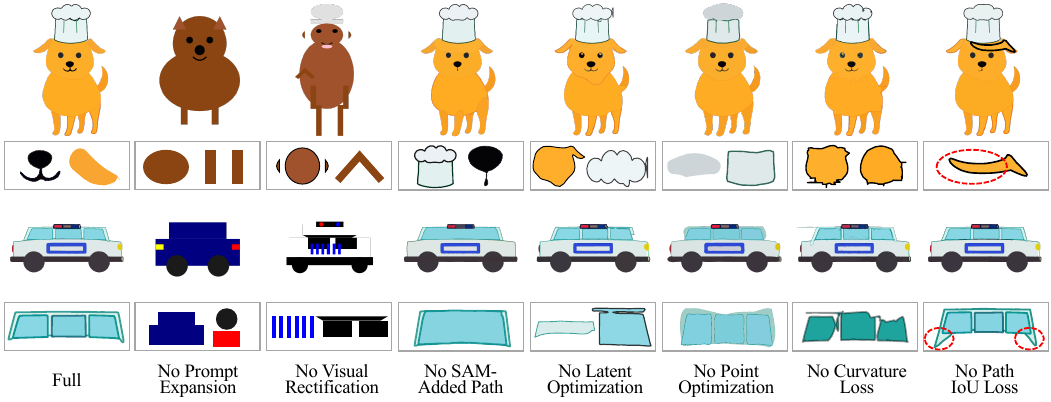}
    \caption{%
        Qualitative results of ablation study.
    }
    \label{fig:supp_ablation}
\end{figure*}

\textbf{SVG Template}.
We demonstrate the critical role of each component in SVG template generation, with results shown in Figure~\ref{fig:supp_ablation}. 
(1) The prompt expansion phase is crucial for synthesizing complete SVG templates by enriching the initial brief description with essential details and components. Without it, key visual elements are missing, such as the dog's chef hat and the police car's flashing lights, leading to incomplete and semantically deficient outputs. This demonstrates how prompt expansion helps capture essential visual elements needed for a coherent design.
(2) Without visual rectification, the generated SVG contains significant visual inconsistencies that impact both local and global coherence. At the local level, we observe misalignments between connected components (like the dog's ear and head), while at the global level, certain elements become unrecognizable (such as the police car's body) due to improper spatial relationships and proportions. These issues highlight how visual rectification serves as a crucial quality control step in ensuring the generated SVG maintains proper visual structure and semantic clarity.

\noindent{\textbf{Detail Enhancement}}.
The SAM-guided path addition enhances the visual richness of the optimized SVG by incorporating decorative elements and fine details.
Without this step, the generated SVG appears simpler.
As shown in the fourth column of Figure~\ref{fig:supp_ablation}, compared to the complete method in the first column, decorative elements such as the intricate patterns on the dog's chef hat and the detailed windows of the police car are absent, resulting in reduced visual sophistication and artistic appeal.

\noindent{\textbf{SVG Optimization}}.
Our experiments validate the importance of each optimization stage:
(1) Without latent optimization, relying solely on point optimization produces shapes with undesirable artifacts like self-intersections in the dog's hat and distorted corners in the car's window;
(2) Conversely, omitting point optimization yields excessively smooth contours in elements like the hat and window, resulting in a lack of geometric precision.
We further analyze the impact of individual loss terms. The curvature loss plays a vital role in maintaining smooth and natural path contours - when excluded, the optimization generates shapes with noticeably irregular boundaries.
Similarly, the path IoU loss is crucial for spatial consistency, anchoring elements in their designated positions. Without this constraint, components like the dog's ear and window frame drift from their intended locations, compromising the overall compositional integrity.

\section{LLM Prompts}
\label{sec:supp_prompt}
In this section, we present the detailed prompts used in our \sysName.
First, we introduce the system prompt (Table~\ref{tab:supp_system_prompt}), which provides global guidance to the LLM.
We then list three prompts used to create an SVG template: prompt expansion (Table~\ref{tab:supp_prompt_expansion}), SVG script generation (Table~\ref{tab:supp_script_generation}), and visual rectification (Table~\ref{tab:supp_visual_rectification}).
Additionally, we include the prompt that guides the LLM to perform accurate editing based on natural language descriptions (Table~\ref{tab:supp_editing}).

\section{More Results}
\label{sec:supp_results}
To provide a more user-friendly visualization, we have put all generated SVG, including the text-guided generation and editing results, on the project website. Please visit the \url{https://chat2svg.github.io/} to view more outputs.

\clearpage
\onecolumn

\begin{longtable}{p{15.5cm}}
\hline
\cellcolor{green!15} System Prompt \\
\hline
    You are a vector graphics designer tasked with creating Scalable Vector Graphics (SVG) from a given text prompt.

    Your tasks:
    \begin{itemize}
        \item Task 1: **Expand Text Prompt**. The provided prompt is simple, concise and abstract. The first task is imagining what will appear in the image, making it more detailed.
        \item Task 2: **Write SVG Code**. Using the expanded prompt as a guide, translate it into SVG code.
        \item Task 3: **Code Improvement**. Although the SVG code may align with the text prompt, the rendered image could reveal oddities from human perception. Adjust the SVG code to correct these visual oddities.
    \end{itemize}
    
    Constraints:
        \begin{enumerate}
        \item SVG Elements: Use only the specified elements: \texttt{rect}, \texttt{circle}, \texttt{ellipse}, \texttt{line}, \texttt{polyline}, \texttt{polygon}, and short \texttt{path} (up to 5 commands).
        \item Canvas Details: The SVG canvas is defined by a 512$\times$512 unit viewBox. Coordinates start at (0, 0) in the top-left and extend to (512, 512) at the bottom-right.
        \item Element Stacking Order: The sequencing of SVG elements matters; elements defined later in the code will overlap earlier ones.
        \item Colors: Use hexadecimal color values (e.g., \#FF0000). For layers fully enclosed by others, differentiate with distinct colors.
        \item Simplicity: Keep the SVG code simple and clear.
        \item Realism: While using simple shapes, strive to create recognizable and proportionate representations of objects.
        \end{enumerate}

    Note: The SVG you create will serve as an initial draft using simple shapes rather than a fully polished final product with complex paths. Focus on creating a recognizable representation of the prompt using basic geometric forms. \\
\hline
\caption{System Prompt for SVG Template Generation} \label{tab:supp_system_prompt} \\
\end{longtable}

\clearpage

\begin{longtable}{p{15.5cm}}
\hline
\cellcolor{green!15} Prompt Expansion \\
\hline
    Expand the given text prompt to detail the abstract concept. Follow these steps in your response:
    
    \begin{enumerate}
        \item **Expand the Short Text Prompt**. Start by expanding the short text prompt into a more detailed description of the scene. Focus on talking about what objects appear in the scene, rather than merely talking the abstract idea. For example, for the short prompt ``A spaceship flying in the sky'', expand it to ``A silver spaceship soars through the galaxy, with distant planets and shimmering stars forming a vivid backdrop''.
        \item **Object Breakdown and Component Analysis**.
        \begin{enumerate}
            \item For every object in the expanded prompt, add more details to describe the object. You can include color, size, shape, motion, status, or any other relevant details. For example, ``A silver spaceship'' can be expanded into ``A silver spaceship with two large wings, ejecting flames from its thrusters''.
            \item Then, break down each object into its individual components. For instance, the spaceship's components could be ``a body (rectangle), two triangular wings (polygon), a window (circle), and flames (polyline) emitting from the rear thrusters (rectangle)''. You need to list **ALL** parts of each object. If you ignore any part, the system will assume it's not present in the scene. When listing components, explain how each component can be depicted using the specified SVG elements.
        \end{enumerate}
        \item **Scene Layout and Composition**. Propose a logical and visually appealing layout for the final scene. For each object and each component, describe their positions on the canvas, relative/absolute sizes, colors, and relative spatial arrangement (e.g., the hand is connected to the arm, the moon is behind the mountain).
    \end{enumerate} \\
    
    When expanding the prompt, follow these guidelines:
    \begin{enumerate}
        \item When expanding the short text prompt, avoid adding excessive new objects to the scene. Introduce additional objects only if they enhance the completeness of the scene. If the prompt mentions just a single object, you can choose to not introduce new objects and focus instead on enriching the description with more details about that object.
        \item When add details to describe objects, the description can be detailed and vivid, but the language should be clear and concise. Avoid overly complex or ambiguous descriptions.
        \item When breaking down objects into individual components, ensure you list all essential parts typically comprising that object, even if they are not explicitly mentioned in the initial object description.
    \end{enumerate} \\
    
    A Unicorn Example:
    
    \#\#\#\#\#

    Input Text Prompt: ``A unicorn is eating a carrot.'' \newline

    Expanded Prompt:
    \begin{enumerate}
        \item Scene Description: ``The pink unicorn is standing in side view. The unicorn's mouth is open, bitting an orange carrot.''
        \item Object Detail:
        \begin{itemize}
            \item Unicorn: The unicorn, standing in side view, has a pink body and a yellow horn. Its brown tail swings, while its four legs stand still. It has a pink head and a small black eye.
            \item Carrot: The carrot is orange with two green leaves at the top.
        \end{itemize}
        \item Component Breakdown:
        \begin{itemize}
            \item Unicorn: horizontal pink body (ellipse), pink head (ellipse), thin neck (rectangle), four vertical legs (rectangles), brown curved tail (polyline), yello triangle horn atop the head (polygon), round eye (circle), mouth on the head (path)
            \item Carrot: elongated orange triangle body (triangle), two small green triangular leaves (triangle)
        \end{itemize}
    \end{enumerate} \\
    
    Key Components Layout:
    
    \begin{enumerate}
    \item Unicorn
        \begin{enumerate}
            \item Body: An ellipse centered around (256, 256) with rx=90 and ry=60. The body is pink.
            \item Head: An ellipse centered at (342, 166), 30 units wide and 25 units high, oriented to suggest the unicorn's gaze forward.
            \item Neck: A rectangle, 50 units long and 10 units wide, positioned at (312, 168) connecting the head and the body.
            \item Legs: Four rectangles, each 10 units wide and 80 units tall, positioned at (185, 296), (220, 311), (293, 307), and (316, 296) to suggest stability.
            \item Tail: A polyline starting from the back of the body ellipse at (168, 258) and curving to points (122, 298) and (142, 252) to suggest a flowing tail.
            \item Horn: A polygon with points at (331, 140), (336, 110), and (341, 140) to represent the unicorn's triangular horn. The horn is yellow.
            \item Eye: A small circle with a radius of 5 units, placed at (352, 164) on the head.
            \item Mouth: A small curved line positioned at (342, 178) on the head
        \end{enumerate}
    
    \item Carrot
        \begin{enumerate}
            \item Body: An elongated triangle with points at (369, 180), (348, 200), and (356, 172). The carrot body is orange (\#FFA500).
            \item Leaves: Two small triangles positioned at the top of the carrot body, centered around (363, 174). The leaves are green (\#00FF00).
        \end{enumerate}
    \end{enumerate}
    
    \#\#\#\#\#

    Refer to the Unicorn example for response guidance and formatting. Avoiding any unnecessary dialogue in your response.
    
    Here is the text prompt: TEXT\_PROMPT \\
\hline
\caption{Detailed Guidance for the Prompt Expansion Stage} \label{tab:supp_prompt_expansion} \\
\end{longtable}

\newpage

\begin{longtable}{p{15.5cm}}
\hline
\cellcolor{green!15} SVG Script Generation \\
\hline
Write the SVG code following the expanded prompt and layout of key components, adhering to these rules:
\begin{enumerate}
    \item SVG Elements: Use only the specified elements: \texttt{rect}, \texttt{circle}, \texttt{ellipse}, \texttt{line}, \texttt{polyline}, \texttt{polygon}, and short \texttt{path} (up to 5 commands). Other elements like \texttt{text}, \texttt{Gradient}, \texttt{clipPath}, etc., are not allowed. If there is \texttt{path}, the final command should be \texttt{Z}. If a path only contains a single command, you need to increase the \texttt{stroke-width} to make it visible.
    \item Viewbox: The viewbox should be 512 by 512.
    \item Stacking Order: Elements defined later will overlap earlier ones. So if there is a background, it should be defined first.
    \item Colors: Use hexadecimal color values (e.g., \#FF0000). For layers fully enclosed by others, differentiate with distinct colors.
    \item Comments: Include concise phrase to explain the semantic meaning of each element.
    \item Shape IDs: **Every** shape element should have a unique ``id'' starting with ``path\_num''.
\end{enumerate} \\

The translation from the Unicorn's expanded prompt to SVG code is provided below:

\#\#\#\#\#
\begin{lstlisting}[
    language=XML,
    basicstyle=\ttfamily\small,
    showspaces=false,
    showstringspaces=false,
    keepspaces=true,
    columns=flexible,
    breaklines=true,
    breakatwhitespace=true,
    linewidth=14.5cm,  % slightly less than the p{15.5cm} to account for margins
    postbreak=\mbox{\textcolor{red}{$\hookrightarrow$}\space}  % optional: shows where lines break
]
<svg viewBox="0 0 512 512" xmlns="http://www.w3.org/2000/svg">
    <!-- Body -->
    <ellipse id="path_1" cx="256" cy="256" rx="90" ry="60" fill="#ffc0cb"/>
    <!-- Legs -->
    <rect id="path_2" x="185" y="296" width="10" height="80" fill="#d3d3d3"/>
    <rect id="path_3" x="220" y="311" width="10" height="80" fill="#d3d3d3"/>
    <rect id="path_4" x="293" y="307" width="10" height="80" fill="#d3d3d3"/>
    <rect id="path_5" x="316" y="296" width="10" height="80" fill="#d3d3d3"/>
    <!-- Neck -->
    <rect id="path_6" x="312" y="168" width="10" height="50" fill="#ffc0cb"/>
    <!-- Head -->
    <ellipse id="path_7" cx="342" cy="166" rx="30" ry="25" fill="#ffc0cb"/>
    <!-- Eye -->
    <circle id="path_8" cx="352" cy="164" r="5" fill="#000000"/>
    <!-- Tail -->
    <polyline id="path_9" points="168,258 122,298 142,252" fill="none" stroke="#a52a2a" stroke-width="8"/>
    <!-- Horn -->
    <polygon id="path_10" points="331,140 336,110 341,140" fill="#ffff00"/>
    <!-- Unicorn mouth -->
    <path id="path_11" d="M 337 178 Q 342 183 347 178" fill="none" stroke="#000000" stroke-width="2"/>
    <!-- Carrot body -->
    <polygon id="path_12" points="369 180 348 200 356 172 369 180" fill="#ffa500"/>
    <!-- Carrot leaves -->
    <polygon id="path_13" points="363 174 364 163 373 168 363 174" fill="#00ff00"/>
    <polygon id="path_14" points="363 174 356 166 375 173 363 174" fill="#00ff00"/>
</svg>
\end{lstlisting}
\#\#\#\#\# \\

In your answer, avoid any unnecessary dialogue, and include the SVG code in the following format:
\begin{lstlisting}[
    language=XML,
    basicstyle=\ttfamily\small,
    showspaces=false,
    showstringspaces=false,
    keepspaces=true,
    columns=flexible,
    breaklines=true,
    breakatwhitespace=true,
    linewidth=14.5cm,  % slightly less than the p{15.5cm} to account for margins
    postbreak=\mbox{\textcolor{red}{$\hookrightarrow$}\space}  % optional: shows where lines break
]
```svg
  svg_code
```
\end{lstlisting}\\
\hline
\caption{Detailed Prompt for the SVG Script Generation Stage} \label{tab:supp_script_generation} \\
\end{longtable}

\begin{longtable}{p{15.5cm}}
\hline
\cellcolor{green!15} Visual Rectification \\
\hline

The SVG code you provide might have a critical issue: while it adheres to the text prompt, the rendered image could reveal real-world inconsistencies or visual oddities. For example:
\begin{enumerate}
    \item Misalignments: The unicorn's legs may appear detached from the body.
    \item Hidden elements: The snowman's arms could be hidden if they blend with the body due to identical colors and overlapping elements, making them indistinguishable.
    \item Unrecognizable object: The SVG code includes a tiger, but the rendered image is unrecognizable due to a disorganized arrangement of shapes.
    \item Disproportionate scaling: The squirrel's tail might appear overly small compared to its body.
    \item Color: If a shape is purely white and placed on a white background, it may seem invisible in the final image. If there is no background, try to avoid using white for the shape.
    \item Incorrect path order: Incorrect path order can cause unintended overlaps, such as the face being completely covered by the hair or hat.
\end{enumerate} \\

These issues may not be evident in the SVG code but become apparent in the rendered image. \newline

The provided image is rendered from your SVG code. You need to do the following:
\begin{enumerate}
    \item First, carefully examine the image and SVG code to detect visual problems. Please list ALL the visual problems you find. If the image is severely flawed/unrecognizable, consider rewriting the entire SVG code.
    \item Second, adjust the SVG code to correct these visual oddities, ensuring the final image appears more realistic and matches the expanded prompt.
    \item When adding/deleting/modifying elements, ensure the IDs are unique and continuous, starting from ``path\_1'', ``path\_2'', etc. For example, if you delete ``path\_3'', rename ``path\_4'' to ``path\_3'' and ``path\_5'' to ``path\_4'' to maintain continuity.
\end{enumerate} \\

Your task is NOT to modify the SVG code to better match the image content, but to identify visual oddities in the image and suggest adjustments to the SVG code to correct them. You are not permitted to delete any SVG elements unless rewriting is involved. \newline

In your answer, include the SVG code in the following format:
\begin{lstlisting}[
    language=XML,
    basicstyle=\ttfamily\small,
    showspaces=false,
    showstringspaces=false,
    keepspaces=true,
    columns=flexible,
    breaklines=true,
    breakatwhitespace=true,
    linewidth=14.5cm,  % slightly less than the p{15.5cm} to account for margins
    postbreak=\mbox{\textcolor{red}{$\hookrightarrow$}\space}  % optional: shows where lines break
]
```svg
  svg_code
```
\end{lstlisting} \\
\hline
\caption{Detailed Prompt for the Visual Rectification Stage} \label{tab:supp_visual_rectification} \\
\end{longtable}

\newpage

\begin{longtable}{p{15.5cm}}
\hline
\cellcolor{green!15} Editing \\
\hline
Analyze the given editing prompt to understand the user's intention. Classify the editing instruction into one of (or a combination of) the following categories:
\begin{enumerate}
    \item **Object Addition**: Adding a new object to the scene.
    \item **Object Removal**: Removing an existing object from the scene.
    \item **Object Modification**: Changing an existing object in the scene (e.g., color, size, position, pose, layout).
\end{enumerate} \\

Follow these steps based on the type of editing instruction:
\begin{itemize}
    \item **Object Addition**:
    \begin{enumerate}
        \item Detailed Description: For each new object, add more details to describe the object. You can include color, size, shape, motion, status, or any other relevant details. For example, ``A silver spaceship'' can be expanded into ``A silver spaceship with two large wings, ejecting flames from its thrusters''.
        \item Component Breakdown: Break down each object into its individual components. For instance, the spaceship's components could be ``a body (rectangle), two triangular wings (polygon), a window (circle), and flames (polyline) emitting from the rear thrusters (rectangle)''. You need to list **ALL** parts of each object. If you ignore any part, the system will assume it's not present in the scene. When listing components, explain how each component can be depicted using the specified SVG elements.
        \item Global Layout: Propose a global layout for each new object, i.e., describing its spatial relationship to existing elements.
        \item Local Components Layout: Propose a local components layout, describing arrangement of its components, including their relative sizes and positions.
        \item Stacking Order: Specify the layering order of new elements, especially for overlapping objects, to ensure the correct visual effect.
    \end{enumerate}
    \item **Object Removal**:
    \begin{enumerate}
        \item Identify the object(s) to be removed with precise descriptions.
        \item Specify any adjustments needed for remaining elements to maintain scene coherence.
    \end{enumerate}
    \item **Object Modification**:
    \begin{enumerate}
        \item Identify the specific part(s) of the object that need to be changed.
        \item Describe modifications in detail, including exact size, colors (in hexadecimal), and positions where applicable.
        \item For complex modifications, consider treating them as a combination of object removal and addition.
    \end{enumerate}
\end{itemize} \\

Guidelines for expanding the prompt:
\begin{enumerate}
    \item When add details to describe objects, the description can be detailed and vivid, but the language should be clear and concise. Avoid overly complex or ambiguous descriptions.
    \item When breaking down objects into individual components, ensure you list all essential parts typically comprising that object, even if they are not explicitly mentioned in the initial object description.
    \item For object modifications, provide exact specifications (e.g., ``Increase the unicorn's horn length by 20 units'').
    \item Consider the overall composition and balance of the scene when adding or modifying elements.
\end{enumerate} \\

Operation Summary:

In the original SVG code, each shape has an id attribute. After generating the edited SVG code, summarize the operations performed in the following format:
\begin{enumerate}
    \item Element Modification: [id\_1, id\_2, ...] (for elements that were modified but kept the same stacking order)
    \item Element Removal: [id\_1, id\_2, ...] (for elements that were removed)
    \item Element Addition: start\_path\_id, [id\_1, id\_2, ...] (for newly added elements; start\_path\_id is the id of the path in the original SVG code, indicating that new elements should be inserted after the element with id start\_path\_id. If inserting at the beginning, set start\_path\_id to an empty string ``''.)
\end{enumerate}
Note: The ids in the lists refer to the ids from the original SVG code, not the edited SVG code. \\
\hline
\caption{Detailed Prompt for the Editing Stage} \label{tab:supp_editing} \\
\end{longtable}

\clearpage
\twocolumn